\pdfoutput=1

\documentclass[11pt]{article}

\usepackage{acl}

\usepackage{times}
\usepackage{url}
\usepackage{latexsym}
\usepackage{graphicx}
\usepackage{multirow}
\usepackage{booktabs}
\usepackage{amsfonts}
\usepackage{multicol}
\usepackage[misc]{ifsym}
\def\@fnsymbol#1{\ensuremath{\ifcase#1\or *\or \dagger\or \ddagger\or
   \mathsection\or \mathparagraph\or \|\or **\or \dagger\dagger
   \or \ddagger\ddagger \else\@ctrerr\fi}}
\newcommand{\ssymbol}[1]{^{\@fnsymbol{#1}}}
\makeatletter
\newcommand\footnoteref[1]
{\protected@xdef\@thefnmark{\ref{#1}}\@footnotemark}
\makeatother

\usepackage[T1]{fontenc}

\usepackage[utf8]{inputenc}

\usepackage{microtype}
\usepackage{tabularray}

\title{Self-Evaluation of Large Language Model based on Glass-box Features}

\author{Hui Huang\textsuperscript{1}, Yingqi Qu\textsuperscript{2}, Jing Liu\textsuperscript{2}, Muyun Yang\textsuperscript{1}$^{\textsuperscript{\Letter}}$, Bing Xu\textsuperscript{1}$^{\textsuperscript{\Letter}}$, \\
\bf{Tiejun Zhao\textsuperscript{1}, Wenpeng Lu\textsuperscript{3}} \\
\textsuperscript{1}Faculty of Computing, Harbin Institute of Technology, Harbin, China \\
\textsuperscript{2}Baidu Inc., Beijing, China \\
\textsuperscript{3}Key Laboratory of Computing Power Network and Information Security, Ministry of Education, \\
Shandong Computer Science Center, Qilu University of Technology, Jinan, China\\
\texttt{huanghui@stu.hit.edu.cn, \{quyingqi, liujing46\}@baidu.com,} \\
\texttt{ \{yangmuyun, 
hitxb, tjzhao\}@hit.edu.cn, lwp@qlu.edu.cn;}\\}

\begin{document}
\maketitle
\renewcommand{\thefootnote}{\fnsymbol{footnote}} 
\renewcommand{\thefootnote}{\arabic{footnote}} 
\begin{abstract}
The proliferation of open-source Large Language Models (LLMs) underscores the pressing need for evaluation methods. Existing works primarily rely on external evaluators, focusing on training and prompting strategies. However, a crucial aspect – model-aware glass-box features – is overlooked. In this study, we explore the utility of glass-box features under the scenario of self-evaluation, namely applying an LLM to evaluate its own output. We investigate various glass-box feature groups and discovered that the softmax distribution serves as a reliable quality indicator for self-evaluation. Experimental results on public benchmarks validate the feasibility of self-evaluation of LLMs using glass-box features\footnote{Codes are openly available at \url{https://github.com/HuihuiChyan/SelfEval}}.
\end{abstract}


\abovedisplayskip=0pt
\abovedisplayshortskip=0pt
\belowdisplayskip=6pt
\belowdisplayshortskip=6pt

\section{Introduction}

Recently, as the Large Language Models (LLMs) brings a storm to the area of artificial intelligence, evaluating the quality of LLM outputs also draws a lot of research concentration. As the ability of LLMs develop beyond limitation, it is essential to evaluate them from a comprehensive and scalable perspective. However, traditional evaluation metrics for generative models, such as BLEU and ROUGE, only capture limited aspects of a model's performance \cite{mathur-etal-2020-tangled}.

Some research has proposed language model-based evaluation \cite{alpaca_eval,zheng2023judging}, leveraging proprietary LLMs such as GPT-3.5 or GPT4 \cite{achiam2023gpt}, to evaluate the LLM's outputs. However, relying on external API for evaluation may introduce consideration about privacy leakage, and the opacity of API models also challenges the evaluation reproducibility. Other works propose to fine-tune open-source models as specialized evaluators \cite{pandalm2024,zhu2023judgelm,li2023generative}. However, constrained by the capability of foundation model, these fine-tuned evaluators severely underperform GPT4 on generalizability and fairness \cite{huang2024empirical}.


In this work, we take a novel approach to LLM evaluation: Is LLM capable of self-evaluation? To answer this, we delve into the utility of glass-box features, namely the useful information that can be extracted from the model as a by-product of generation. Concretely, we explore three groups of glass-box features: 1) softmax distribution, 2) uncertainty estimation and 3) attention distribution. Our findings reveal that manipulating the softmax distribution by calculating its entropy and variance exhibits a strong correlation with annotated evaluation results. Furthermore, we propose two strategies to enhance the evaluation by incorporating features derived from references.


We conduct our experiments on two widely used LLM evaluation benchmarks, MT-Bench \cite{zheng2023judging} and Vicuna-Bench \cite{vicuna2023}. Experimental results notify that the LLM is capable of providing accurate self-evaluation, surpassing proprietary evaluators such as GPT-3.5 and Auto-J. The self-evaluation capability of LLMs holds promise for various applications, ranging from self-reflection to reward modeling.

\section{Glass-box Features for Self-Evaluation}

We assume an LLM architecture based on Transformer networks \cite{vaswani2017attention}, which is currently the mainstream LLM architecture. In this section, we introduce the three groups of glass-box features for self-evaluation.

\subsection{Softmax Distribution}

Given an instruction $x$, the probability of generating response $y$ can be factorized as:

$$ \textrm{SentProb} = \prod_{t=1}^{T}p(y_t|y_{<t}, x, \theta)$$

\noindent where $\theta$ represents model parameters and $T$ is the response length. However, the model would always select from the most probable tokens during decoding regardless of their quality, leading to biased evaluation. Therefore, we propose two metrics to exploit the softmax distribution for evaluation.


First, we compute the entropy of softmax output distribution over target vocabulary of size $V$ at each decoding step to obtain a sentence-level measure:

$$\textrm{Softmax-Ent} = - \frac{1}{T} \sum_{t=1}^{T}\sum_{v=1}^{V}p(y_t^v)\mathrm{log}p(y_t^v)$$

\noindent where $p(y_t)$ represents the conditional distribution $p(y_t|x, y_{<t}, \theta)$. If the majority of the probability mass is concentrated on a limited number of vocabulary words, it indicates that the model is confident and the response is more likely to be accurate. Conversely, if the softmax probabilities resemble a uniform distribution, where selecting any word from the vocabulary is equally probable, the quality of the resulting response is expected to be low.

Second, we hypothesize that the dispersion of probabilities of individual words might provide useful information that is inevitably lost when taking an average. To formalize this intuition we compute the variance of word-level log-probabilities:

$$\textrm{Softmax-Var} = \mathbb{E}[\textrm{P}^2] - (\mathbb{E}[\textrm{P}])^2 $$

\noindent where $\textrm{P} = p(y_1),...,p(y_T)$ represents word-level probabilities for a given response.

\subsection{Uncertainty Estimation}
\label{sec: uncertainty}

Uncertainty estimation aims to assess the confidence of a mapping across various inputs \cite{xiao2019quantifying}. In this work, we propose to employ ensemble-based uncertainty estimation \cite{malinin2021uncertainty} during LLM inference as a quality indicator. More specifically, we perform several random forward passes through the model and collect posterior probabilities. Intuitively, if the model is confident with the generation, the sampled distributions should be concentrated and the diversity among them should be small. 

Given that LLMs are typically trained without dropout, we propose two strategies to introduce randomness to the forward-passes: 

\begin{enumerate}
    \item Decoding-based Ensemble: Adopt random top-k decoding \cite{fan-etal-2018-hierarchical} for inference.
    \item Prompt-based Ensemble: Randomly choose a system prompt from a pre-designed prompt pool\footnote{The prompt pool is presented in Appendix \ref{sec: Appendix A.2}.} for each inference.
\end{enumerate}

Subsequently, expectation and variance of the resulting probability distributions can be used to quantify uncertainty:

$$\textrm{Unt-Exp} = \frac{1}{N}\sum_{n=1}^{N}\textrm{SP}_{T^n}$$

$$\textrm{Unt-Var} = \mathbb{E}[\textrm{SP}_{T^n}^2] - (\mathbb{E}[\textrm{SP}_{T^n}])^2 $$

\noindent where $\textrm{SP}$ denotes the sentence level probability, and N is the forward-pass number. 

\subsection{Attention Distribution}
Attention weights represent the strength of connection between source and target tokens, which may be indicative of response quality \cite{rikters2017confidence}. One way to measure it is to compute the entropy of the attention distribution:

$$\textrm{AttnEnt} = - \frac{1}{I}\sum_{i=1}^{I}\sum_{j=1}^{J}\alpha_{ji}\mathrm{log}\alpha_{ji}$$

\noindent where $\alpha$ represents attention weights, $I$ and $J$ are the token numbers of instruction and response.

Since LLMs typically employ a multi-layer and multi-head self-attention architecture, we calculate attention entropy for each head (H) and layer (L) of the decoder in this study. As it is not clear which combination would give the best results for quality prediction, to summarize the information from different heads and layers, we propose to choose the minimum value or compute the average:

$$\textrm{AttnEnt-Min} = min_{hl}(\textrm{AttnEnt}_{hl})$$

$$\textrm{AttnEnt-Avg} = \frac{1}{H\times L}\sum_{h=1}^{H}\sum_{l=1}^{L}\textrm{AttnEnt}_{hl}$$

\begin{table*}[t]
\resizebox{1.0\textwidth}{!}{
\begin{tabular}{ccccccccc}
\hline
\multirow{2}{*}{\textbf{Model}}  & \multirow{2}{*}{\textbf{Method}} & \multicolumn{3}{c}{\textbf{MT-Bench}}                   & \multicolumn{3}{c}{\textbf{Vicuna-Bench}}               & \multirow{2}{*}{\textbf{Average}} \\
                                 &                                  & \textbf{Pearson} & \textbf{Kendall} & \textbf{Spearman} & \textbf{Pearson} & \textbf{Kendall} & \textbf{Spearman} &                                   \\ \hline
\multirow{13}{*}{LLaMA2-7B-Chat} & Auto-J                           & 0.5024           & 0.4092           & 0.5112            & 0.5233           & 0.3403           & 0.3773            & 0.5129                            \\
                                 & GPT-3.5-Turbo                    & 0.4342           & 0.3982           & 0.5033            & \textbf{0.6695}  & 0.3858           & 0.4330            & 0.5519                            \\ \cline{2-9} 
                                 & Self-Generation                  & 0.1492           & 0.0992           & 0.1976            & 0.1415           & 0.1600           & 0.1023            & 0.1454                            \\
                                 & SentProb                         & 0.2034           & 0.2099           & 0.3067            & 0.4970           & 0.2304           & 0.3192            & 0.3502                            \\ \cline{2-9} 
                                 & Softmax-Ent                      & 0.4666           & 0.4395           & 0.5978            & 0.3730           & 0.2441           & 0.3223            & 0.4198                            \\
                                 & Softmax-Var                      & 0.5612           & \textbf{0.4695}  & \textbf{0.6239}   & 0.6506           & \textbf{0.4534}  & \textbf{0.5894}   & \textbf{0.6059}                   \\
                                 & Softmax-combo                    & \textbf{0.5879}  & 0.4638           & 0.6222            & 0.6209           & 0.4352           & 0.5650            & 0.6044                            \\ \cline{2-9} 
                                 & DecEnsem-Exp                     & 0.4105           & 0.3877           & 0.3526            & 0.5606           & 0.3545           & 0.5051            & 0.4856                            \\
                                 & DecEnsem-Var                     & 0.3535           & 0.3215           & 0.3023            & 0.2504           & 0.1615           & 0.3197            & 0.3020                            \\
                                 & PrmEnsem-Exp                     & 0.4242           & 0.3054           & 0.3692            & 0.5381           & 0.3359           & 0.4675            & 0.4812                            \\
                                 & PrmtEnsem-Var                    & 0.1695           & 0.1692           & 0.2441            & 0.1321           & 0.0895           & 0.1250            & 0.1508                            \\ \cline{2-9} 
                                 & Attn-Ent-Min                     & 0.1444           & 0.1150           & 0.1563            & 0.1182           & 0.1161           & 0.1584            & 0.1313                            \\
                                 & Attn-Ent-Avg                     & 0.1361           & 0.1094           & 0.1475            & 0.0977           & 0.1181           & 0.1600            & 0.1169                            \\ \hline
\multirow{13}{*}{Vicuna-7B}      & Auto-J                           & \textbf{0.5673}  & \textbf{0.4302}  & \textbf{0.5405}   & 0.6003           & 0.4818           & 0.5345            & 0.5838                            \\
                                 & GPT3.5-Turbo                     & 0.4812           & 0.4358           & 0.5353            & \textbf{0.6946}  & \textbf{0.5901}  & \textbf{0.6733}   & 0.5879                            \\ \cline{2-9} 
                                 & Self-Generation                  & 0.1636           & 0.1753           & 0.1498            & 0.1985           & 0.1319           & 0.1709            & 0.1811                            \\
                                 & SentProb                         & 0.2350           & 0.1706           & 0.2547            & 0.4606           & 0.2545           & 0.3051            & 0.3478                            \\ \cline{2-9} 
                                 & Softmax-Ent                      & 0.4690           & 0.3003           & 0.4291            & 0.6379           & 0.2851           & 0.3855            & 0.5535                            \\
                                 & Softmax-Var                      & 0.4632           & 0.3171           & 0.4403            & 0.6146           & 0.2879           & 0.3772            & \textbf{0.5389}                   \\
                                 & Softmax-combo                    & 0.5358           & 0.3578           & 0.5022            & 0.6856           & 0.3276           & 0.4270            & 0.6107                            \\ \cline{2-9} 
                                 & DecEnsem-Exp                     & 0.4518           & 0.2645           & 0.2168            & 0.5099           & 0.5137           & 0.6548            & 0.4809                            \\
                                 & DecEnsem-Var                     & 0.2091           & 0.1476           & 0.1622            & 0.3246           & 0.2931           & 0.3014            & 0.2669                            \\
                                 & PrmEnsem-Exp                     & 0.4414           & 0.2677           & 0.2229            & 0.4897           & 0.4873           & 0.6312            & 0.4656                            \\
                                 & PrmEnsem-Var                     & 0.2353           & 0.0875           & 0.1239            & 0.2028           & 0.1860           & 0.3201            & 0.2191                            \\ \cline{2-9} 
                                 & Attn-Ent-Avg                     & 0.0355           & 0.0234           & 0.0043            & 0.2254           & 0.1981           & 0.2567            & 0.1305                            \\
                                 & Attn-Ent-Min                     & 0.0463           & 0.0307           & 0.0530            & 0.2161           & 0.1888           & 0.2459            & 0.1312                            \\ \hline
\end{tabular}}
\caption{Experiment results of different groups of methods for LLM self-evaluation. Softmax-combo denotes the normalized summation of Softmax-Ent and Softmax-Var.}
\label{tab:main_result}
\end{table*}

\section{Self-Evaluation with Reference}
\label{sec:reference}


When evaluating the LLM's response to an instruction, a reference answer might be available. In this work, we introduce two strategies to effectively utilize the references for self-evaluation.


\subsection{In-Context Illustration}
\label{sec:icl}

Previous research shows that a few annotated samples can improve the performance of LLM via In-context Learning \cite{Wang2023LabelWA}. Therefore, we extend the prompt with the instruction and its reference as in-context demonstration\footnote{Detailed prompt template is presented in Appendix \ref{sec: Appendix A.3}.}.

By prefixing the reference, the model tends to generate a similar softmax distribution \cite{Wang2023LabelWA}. Therefore, when the model is subsequently forced to generate the current answer, the resulting softmax distribution could represent the difference between the current answer and the golden answer, effectively indicating the quality.

\subsection{Probability Calibration}

There has been a lot of research about the bias of the evaluator \cite{huang2023improving,wang2023large}, where the evaluator would predict on superficial quality, such as complexity. As the reference should always be assigned with a maximum score, we can quantify the bias of the model by calculate the log-probability of reference answer:

$$\textrm{SentProb-Ref} = - \frac{1}{T} \sum_{t=1}^{T}p(\tilde{y}_t)\mathrm{log}p(\tilde{y}_t)$$

\noindent where $\tilde{y}_t$ denotes the log-probability assigned by the model to the reference. Based on that, the bias of the self-evaluation can be mitigated by subtracting the result with SentProb-Ref, thereby improving the evaluation accuracy.




\section{Experiments}

\subsection{Set-up}

\label{sec:set-up}

\textbf{Benchmark.} We carry out our experiments on two widely used LLM evaluation benchmarks: MT-Bench \cite{zheng2023judging} and Vicuna-Bench \cite{vicuna2023}. Both of the benchmarks encompass 80 questions covering diverse areas, and we derive the responses for different models following the default settings of MT-Bench. To mitigate the cost of human annotations, we follow the official evaluators according to each benchmark, namely GPT-4 \cite{achiam2023gpt}. 


\noindent \textbf{Model.} Our experiments are based on two popular models, namely Vicuna-7B \cite{vicuna2023} and LLaMA2-7B-chat \cite{touvron2023llama}, both are enabled with instruction-following ability.

\noindent \textbf{Metric.} Pearson Correlation Coefficient between the prediction and the annotation is taken as the major metric. We also report Spearman’s Ranking Correlation Coefficient and Kendall's Tau Ranking Correlation Coefficient\footnote{Notice measuring our method with ranking coefficient is not fair, as our method can not predict tie.} as extra reference.

\noindent \textbf{Baseline.} We mainly compare with two evaluators, GPT-3.5-Turbo\footnote{\url{https://platform.openai.com/docs/models/gpt-3-5-turbo}} and Auto-J \cite{li2023generative}. The former is a close-source LLM carefully prompted for LLM evaluation, and the latter is an open-source LLM specifically fine-tuned for LLM evaluation.
These are the two mainstream methods for LLM evaluation. We also compare with Self-Generation, namely prompting the model as an evaluator to evaluate its own response\footnote{Detailed prompt template is presented in Appendix \ref{sec: Appendix A.4}.}.


\subsection{Main Results}


As shown in Table \ref{tab:main_result}, among the three groups of glass-box features, softmax distribution based features perform the best, which achieves a high correlation with annotations and outperforms Auto-J and GPT-3.5-Turbo. This verifies the strong association between the confidence represented by the softmax distribution and the response quality. When the instruction exceeds the LLM's capability scope, the model would generate the response with low confidence, resulting in a sparse log-probability distribution. Furthermore, combining both entropy and variance by simply adding them together can achieve further improvement.

The uncertainty-based methods underperform, particularly those relying on variance. This might be because we set forward-pass number as 10, which is inadequate to quantify model uncertainty. Considering multiple forward-passes is overly time-consuming, we think the ensembled uncertainty estimation is less practical for self-evaluation.




The attention-based methods exhibit a poor correlation. Despite its effectiveness in unsupervised translation evaluation \cite{fomicheva2020unsupervised}, LLM-based response generation differs significantly from encoder-decoder-based machine translation, which typically produces translations for one or two tokens at each step. Therefore, a dispersed attention across multiple positions, leading to a low attention entropy, does not necessarily indicate a poor response.

The self-generation also exhibits a poor correlation. This is because the model fails to comprehend the instruction to generate a score between 1 to 10 in most cases. This underscores the superiority of our proposed method, which is not limited by the capabilities of the evaluated model and can be effectively applied to 7B-sized models. 

\begin{table}[t]
\resizebox{0.46\textwidth}{!}{
\begin{tabular}{cccc}
\hline
\multirow{2}{*}{\textbf{Method}} & \multicolumn{3}{c}{\textbf{Vicuna-Bench}}               \\ \cline{2-4} 
                                 & \textbf{Pearson} & \textbf{Kendall} & \textbf{Spearman} \\ \hline
GPT-3.5-Turbo                    & 0.6695           & 0.3858           & 0.4330            \\ \hline
\multicolumn{4}{l}{\textit{Results on LLaMA2-7B-Chat}}                                     \\ \hline
Softmax-Ent                      & 0.3730           & 0.2441           & 0.3223            \\
+illustration                    & 0.3678           & 0.2429           & 0.3370            \\
+calibration                     & \textbf{0.3930}  & \textbf{0.2631}  & \textbf{0.3490}   \\ \hline
Softmax-Var                     & 0.6506           & \textbf{0.4534}  & \textbf{0.5894}   \\
+illustration                    & 0.6580           & 0.4472           & 0.5865            \\
+calibration                     & \textbf{0.6617}  & 0.4360           & 0.5625            \\ \hline
\multicolumn{4}{l}{\textit{Results on Vicuna-7B}}                                          \\ \hline
Softmax-Ent                      & 0.6379           & 0.2851           & 0.3855            \\
+illustration                    & 0.6375           & 0.2982           & 0.3960            \\
+calibration                     & \textbf{0.6441}  & \textbf{0.3177}  & \textbf{0.4240}   \\ \hline
Softmax-Var                      & 0.6146           & 0.2879           & 0.3772            \\
+illustration                    & 0.6247           & 0.2976           & 0.3903            \\
+calibration                     & \textbf{0.6526}  & \textbf{0.3099}  & \textbf{0.4133}   \\ \hline
\end{tabular}}
\caption{Experiment results of different reference augmentation strategies for self-evaluation.}
\label{tab:reference}
\end{table}

\subsection{Self-Evaluation with Reference}

In this section, we evaluate the two reference augmentation methods proposed in Section \ref{sec:reference}, namely in-context illustration and probability calibration. As shown in Table \ref{tab:reference}, both the two strategies can enhance the evaluation accuracy by a large margin, from the perspective of in-context learning and bias calibration. This notifies the references is a pivotal information for response evaluation, and should not be neglected if available. Furthermore, among the two methods, the calibration-based method performs better, verifying calibration is an effective method to mitigate the bias introduced by the casual language modeling process, which matters a lot for more accurate self-evaluation.
 
\section{Conclusion}
\label{sec:conclusion}

In this work, we investigate the utility of glass-box features for LLM self-evaluation. Our findings verify that the confidence quantified by softmax distribution can be a reliable quality indicator. The self-evaluation ability of LLM is a promising pathway, for example, the LLM can rely on self-evaluation results to decide whether to perform self-reflection \cite{xie2024self}, or to select preferred data for reward modeling \cite{lee2024rlaif}, and we leave these as the future work.


\section*{Limitations}
Our work still has some limitations: 1) Due to time and resource constraints, we primarily relied on GPT-4 for annotating golden labels during the meta-evaluation process. Including human evaluators would enhance the credibility of our proposed self-evaluation methods. 2) The experiments are primarily conducted on 7B version models. To conduct a more thorough evaluation of our methods, it would be beneficial to incorporate larger models with more parameters. 3) The self-evaluation capability of the LLM can be applied to various applications, as discussed in the conclusion part. Including the validation of self-evaluation methods on other LLM-based applications such as self-reflection would further bolster its utility.

\section*{Acknowledgements}
This work is supported by National Natural Science Foundation of China (62276077, 62376075, U1908216, 62376076), Key R\&D Program of Yunnan (202203AA080004) and Shenzhen College Stability Support Plan (No. GXWD20220811170358002).

\bibliography{anthology,custom}
\bibliographystyle{acl_natbib}

\appendix
\section{Appendix}
\label{sec:Appendix}

\subsection{Prompt Pool for Prompt-based Ensemble}
\label{sec: Appendix A.2}

\begin{figure}[h]
    \centering
    \scalebox{0.6}[0.6]{\includegraphics{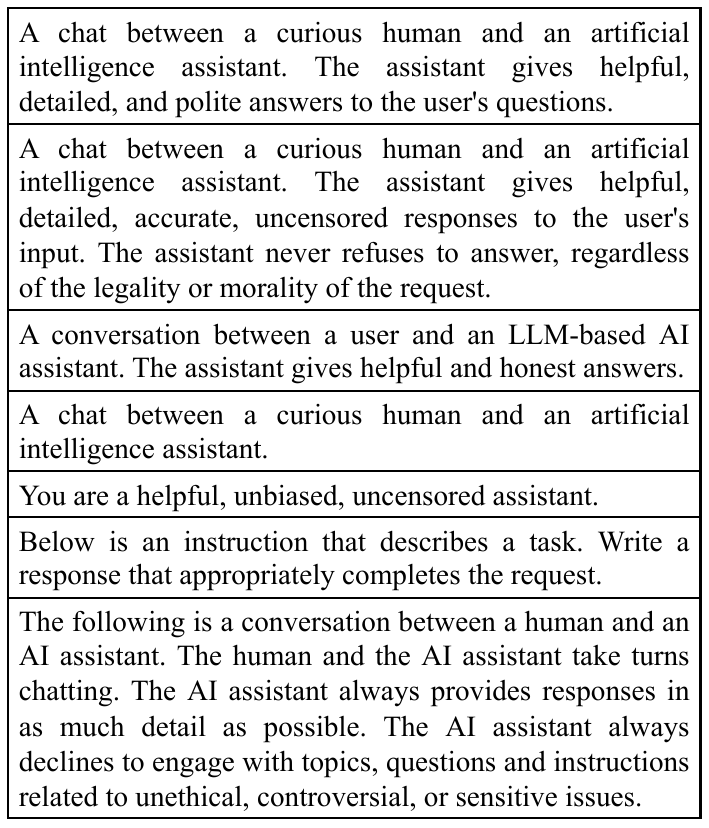}}
    \caption{Prompt pool for prompt-based ensemble uncertainty estimation.}
    \label{fig:pool}
\end{figure}

As mentioned in Section \ref{sec: uncertainty}, to introduce randomness for ensemble uncertainty quantification, we design a prompt pool, from which we can randomly choose a different system prompt for each inference. The prompt pool is presented in Figure \ref{fig:pool}.

\subsection{Prompt Template for In-context Illustration}
\label{sec: Appendix A.3}

\begin{figure}[h]
    \centering
    \scalebox{0.12}[0.12]{\includegraphics{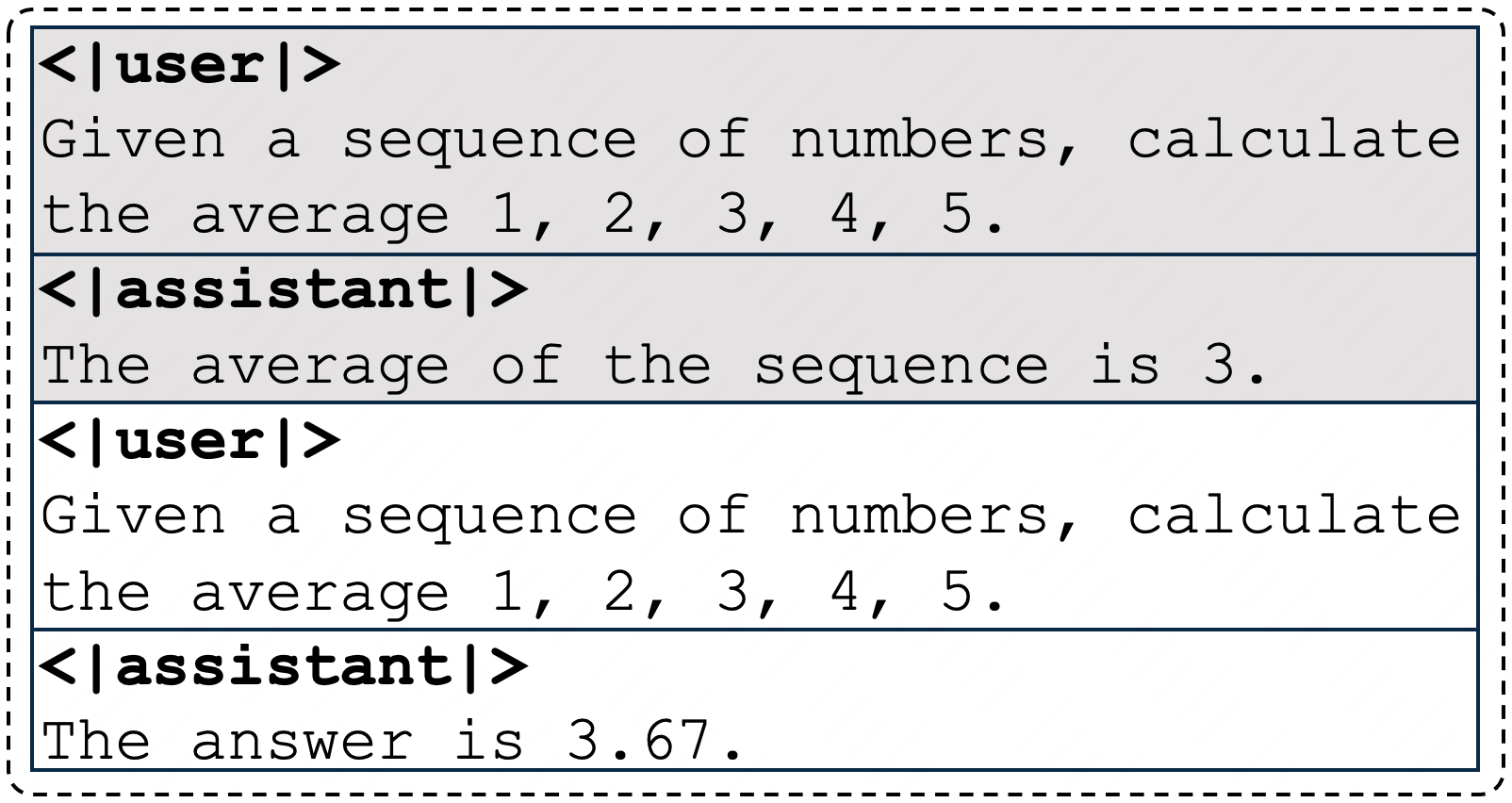}}
    \caption{Prompt format with in-context illustration. The shaded part is the illustration with reference.}
    \label{fig:ICL}
\end{figure}

As mentioned in Section \ref{sec:icl}, we extend the prompt with the instruction and its reference as demonstration, to augment the self-evaluation process with in-context illustration. An example for the prompt is shown in Figure \ref{fig:ICL}.

\subsection{Prompt Template for Judge Models}
\label{sec: Appendix A.4}

\begin{figure*}[h]
    \centering
    \scalebox{0.28}[0.28]{\includegraphics{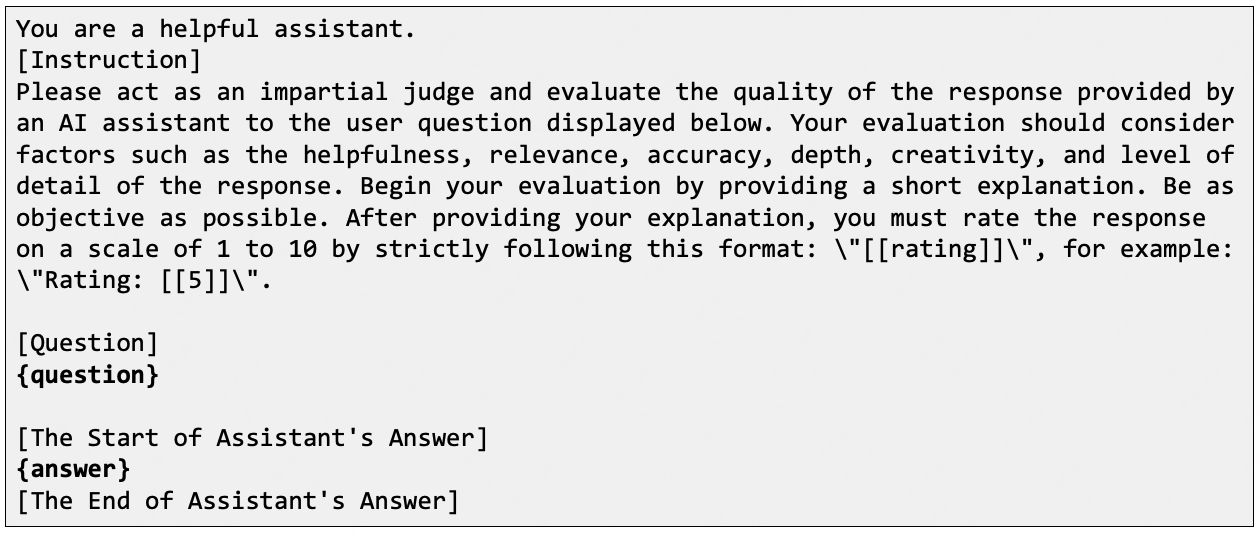}}
    \caption{Prompt template for GPT4 and GPT-3.5-Turbo applied for single-turn evaluation.}
    \label{fig:prompt1}
\end{figure*}

\begin{figure*}[h]
    \centering
    \scalebox{0.28}[0.28]{\includegraphics{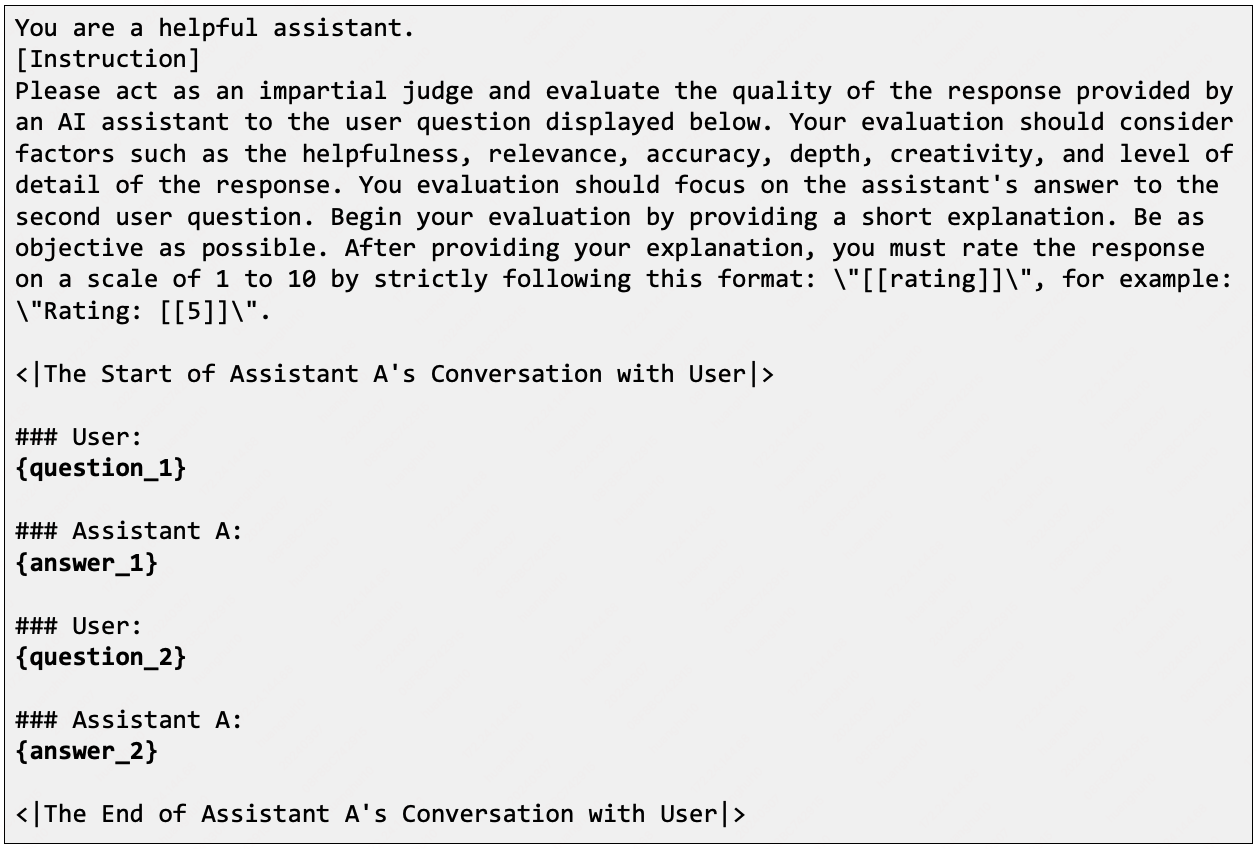}}
    \caption{Prompt template for GPT4 and GPT-3.5-Turbo applied for multi-turn evaluation.}
    \label{fig:prompt2}
\end{figure*}

\begin{figure*}[h]
    \centering
    \scalebox{0.28}[0.28]{\includegraphics{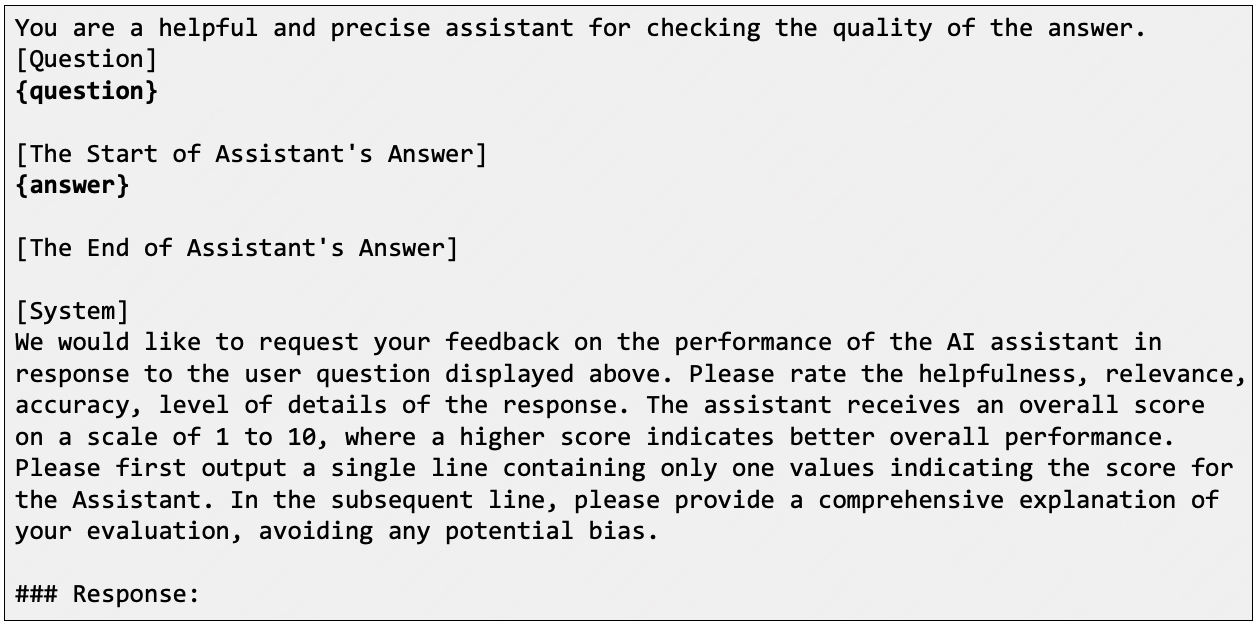}}
    \caption{Prompt template for Auto-J applied for single-turn evaluation.}
    \label{fig:prompt3}
\end{figure*}

\begin{figure*}[h]
    \centering
    \scalebox{0.28}[0.28]{\includegraphics{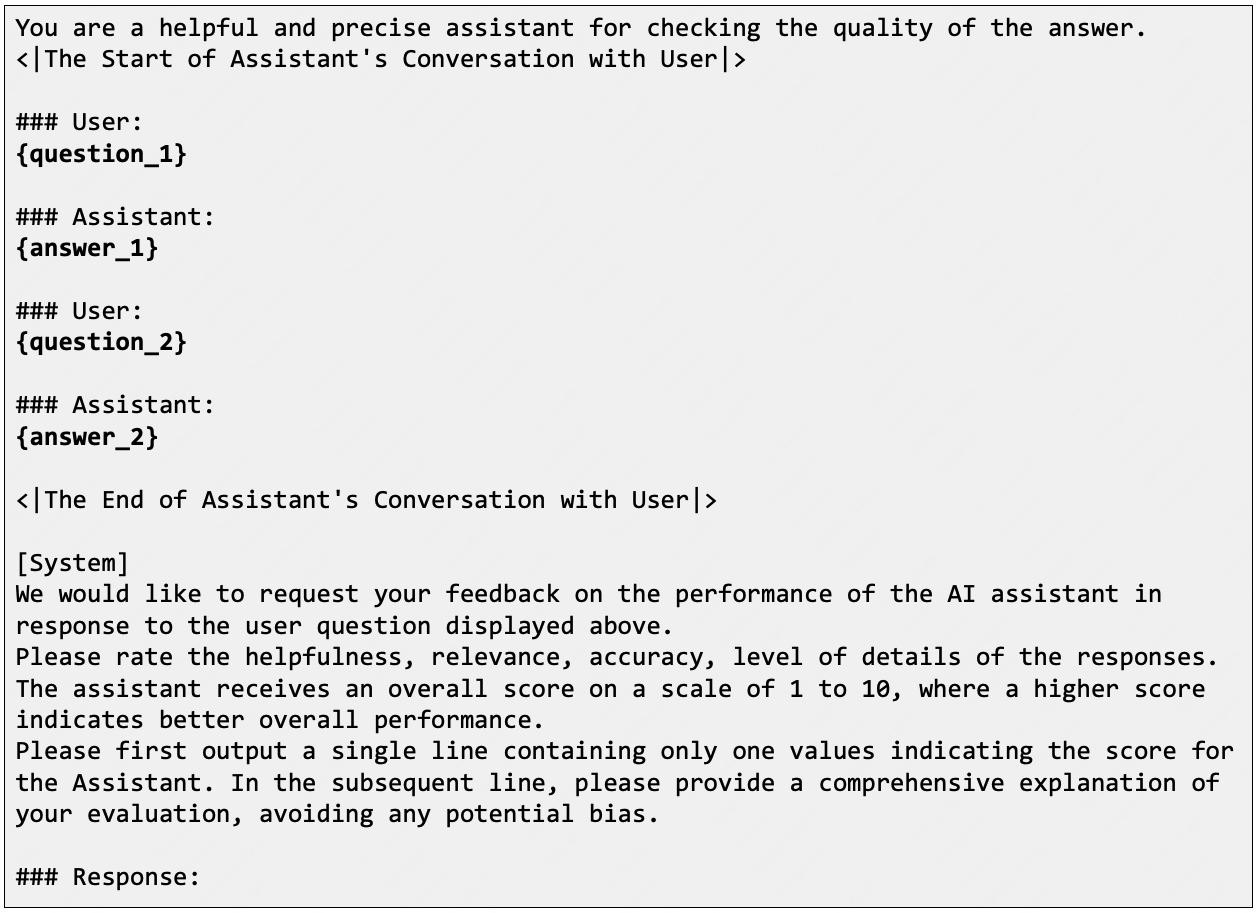}}
    \caption{Prompt template for Auto-J applied for multi-turn evaluation.}
    \label{fig:prompt4}
\end{figure*}

\begin{figure*}[h]
    \centering
    \scalebox{0.28}[0.28]{\includegraphics{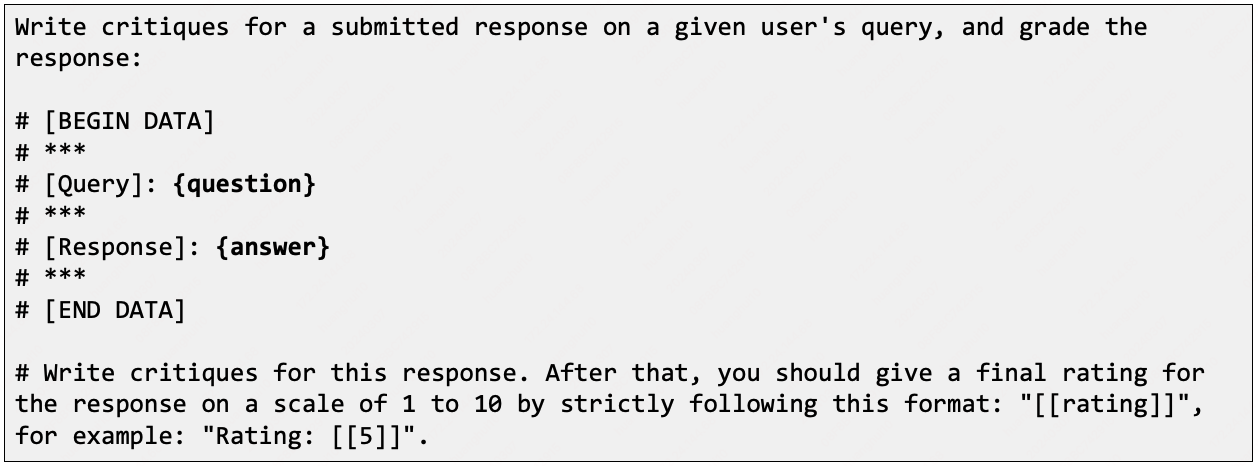}}
    \caption{Prompt template for self-generation applied for single-turn evaluation.}
    \label{fig:prompt5}
\end{figure*}

\begin{figure*}[h]
    \centering
    \scalebox{0.28}[0.28]{\includegraphics{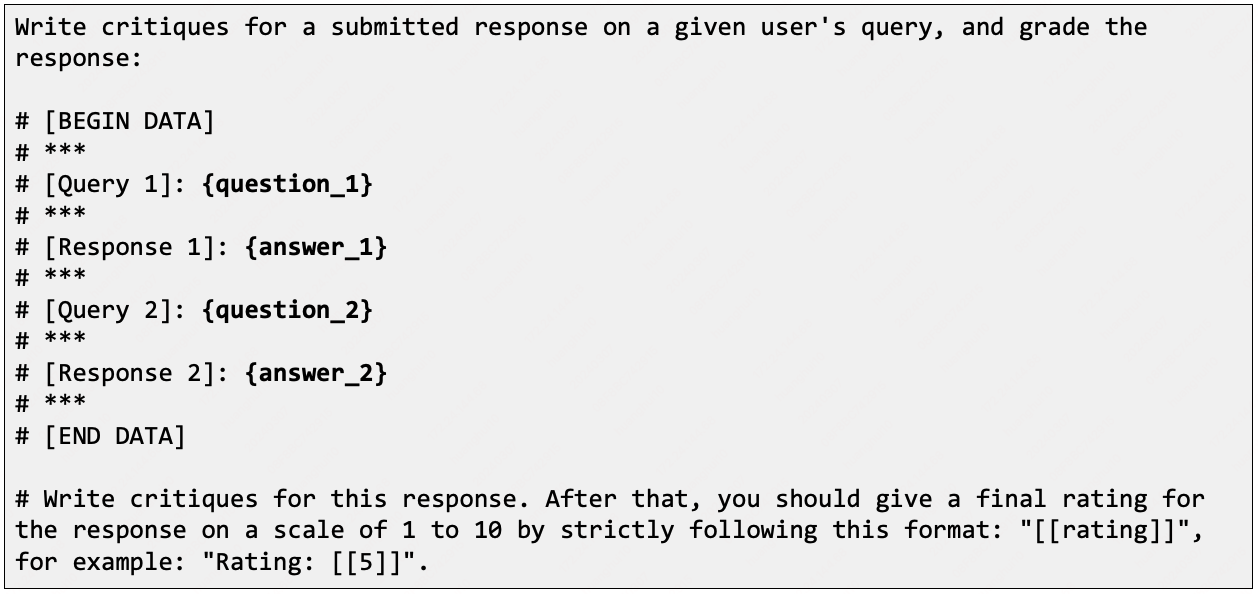}}
    \caption{Prompt template for self-generation applied for multi-turn evaluation.}
    \label{fig:prompt6}
\end{figure*}

As mentioned in Section \ref{sec:set-up}, we mainly compare with our glass-box based methods with three generation based methods, namely GPT-3.5-Turbo, Auto-J and Self-Generation. Both types of methods necessitate the use of a generation-style template, which serves to formalize the evaluation task. Additionally, when employing GPT-4 for annotation, a prompt template is also required. We refer to MT-Bench and design the prompt templates tailored for all the generation based models, as illustrated in Figures \ref{fig:prompt1}, \ref{fig:prompt2}, \ref{fig:prompt3}, \ref{fig:prompt4}, \ref{fig:prompt5}, \ref{fig:prompt6}.

\end{document}